%% file: main.tex
\documentclass[10pt,twocolumn,letterpaper]{article}

\usepackage[accsupp]{axessibility}
\usepackage{iccv}              % To produce the CAMERA-READY version
\makeatletter
\renewcommand{\paragraph}{%
  \@startsection{paragraph}{4}%
  {\z@}{0ex \@plus 1ex \@minus .2ex}{-1em}%
  {\normalfont\normalsize\bfseries}%
}
\makeatother
\newcommand{\Paragraph}[1]{\paragraph{{#1.}}}
\usepackage{graphicx}
\usepackage{amsmath}
\usepackage{amssymb}
\input{preamble}

\definecolor{iccvblue}{rgb}{0.21,0.49,0.74}
\usepackage[pagebackref,breaklinks,colorlinks,allcolors=iccvblue]{hyperref}

\def\paperID{13220} % *** Enter the Paper ID here
\def\confName{ICCV}
\def\confYear{2025}

\title{HAMoBE: Hierarchical and Adaptive Mixture of Biometric Experts for Video-based Person ReID}

\author{
Yiyang Su$^{1}$\thanks{Equal contribution}\quad
Yunping Shi$^{2}$\footnotemark[1]\quad
Feng Liu$^{2}$\quad
Xiaoming Liu$^{1}$\\
$^1$ Department of Computer Science and Engineering, Michigan State University\\
$^2$ Department of Computer Science, Drexel University\\
{\tt\small \{suyiyan1, liuxm\}@msu.edu, \{ys839, fl397\}@drexel.edu}
}

\begin{document}
\maketitle
\input{sec/0_abstract}

\input{sec/1_intro}
\input{sec/2_prior}
\input{sec/3_method}
\input{sec/4_exp}

\input{sec/5_conclusion}

{
    \small
    \bibliographystyle{ieeenat_fullname}
    \bibliography{main}
}

\clearpage
\appendix

\input{sec/X_supp}

\end{document}

%% file: preamble.tex
\usepackage{multirow}
\usepackage{pifont}

%% file: sec/0_abstract.tex
\begin{abstract}
Recently, research interest in person re-identification (ReID) has increasingly focused on video-based scenarios, essential for robust surveillance and security in varied and dynamic environments. However, existing video-based ReID methods often overlook the necessity of identifying and selecting the most discriminative features from both videos in a query-gallery pair for effective matching. To address this issue, we propose a novel Hierarchical and Adaptive Mixture of Biometric Experts (HAMoBE) framework,  which leverages multi-layer features from a pre-trained large model (\emph{e.g.}, CLIP) and is designed to mimic human perceptual mechanisms by independently modeling key biometric features—appearance, static body shape, and dynamic gait—and adaptively integrating them. Specifically, HAMoBE includes two levels: the first level extracts low-level features from multi-layer representations provided by the frozen large model, while the second level consists of specialized experts focusing on long-term, short-term, and temporal features. To ensure robust matching, we introduce a new dual-input decision gating network that dynamically adjusts the contributions of each expert based on their relevance to the input scenarios. Extensive evaluations on benchmarks like MEVID demonstrate that our approach yields significant performance improvements (\eg, +$13.0\%$ Rank1). \href{https://github.com/prevso1088/hamobe}{Project Link}
\end{abstract}

%% file: sec/1_intro.tex
\section{Introduction}\label{sec:intro}

Video-based person re-identification (ReID) is a crucial technology in intelligent systems, aiming to accurately model and match tracklets of the same individual across different camera views~\cite{liu2023farsight,yang2020spatial,mclaughlin2016recurrent,fu2019sta,you2016top,liu2017video,liu2019spatial,liu2025person}. 
ReID is increasingly vital for public security.
The core operation involves processing two video inputs, Video $\#1$ ($\mathbf{V}_1$) and Video $\#2$ ($\mathbf{V}_2$), through a neural network function $\mathcal{E}$ that extracts feature vectors: $f_1=\mathcal{E}(\mathbf{V}_1)$ and $f_2=\mathcal{E}(\mathbf{V}_2)$. These vectors are then compared using a similarity function, where a high value indicates a successful match. 

%---------------------------------------------------------
\begin{figure}
  \centering
  \resizebox{0.98\linewidth}{!}{
  \includegraphics[trim=5mm 0 5mm 0,clip,width=12.0cm]{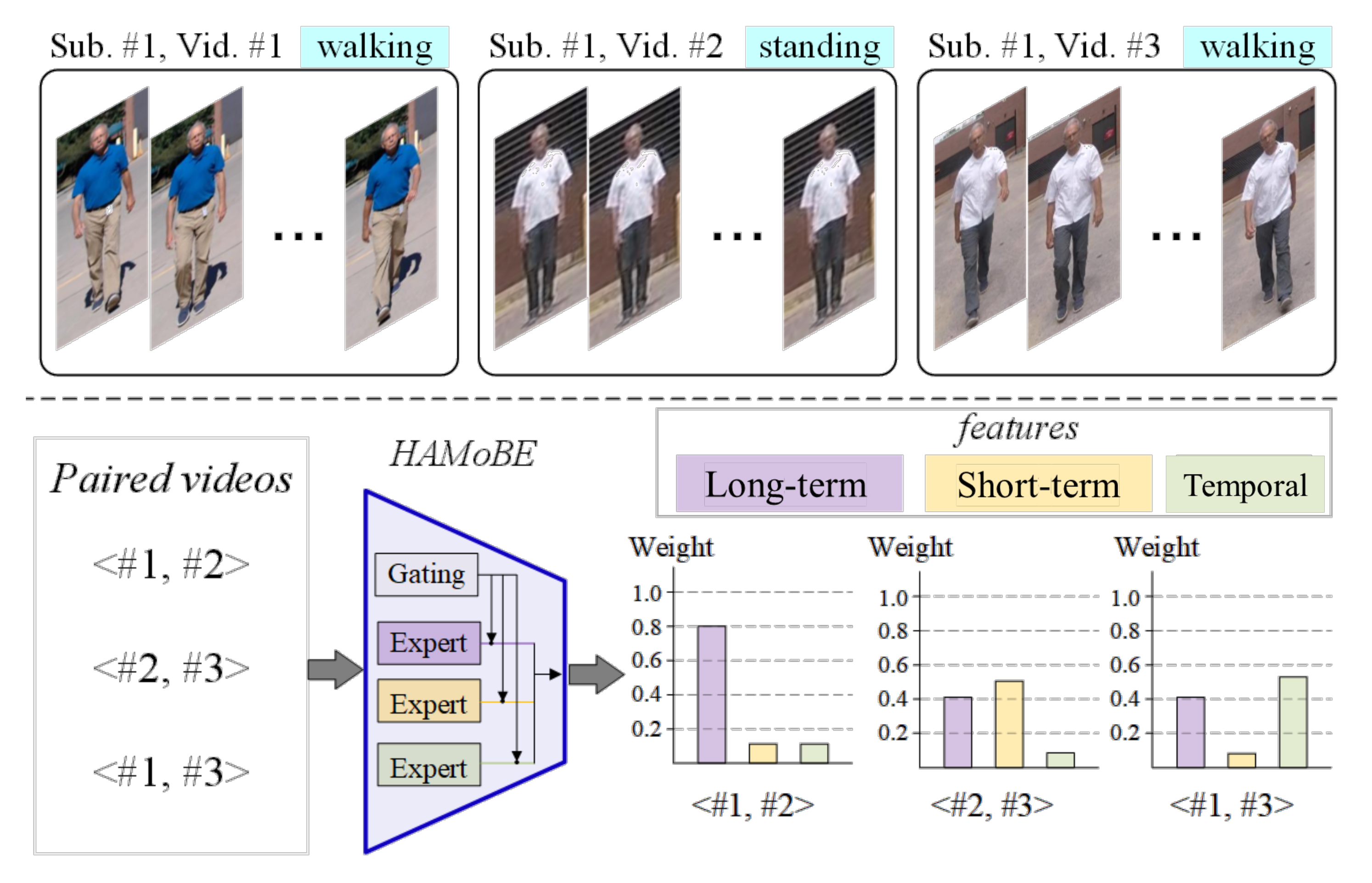}
  }
  \caption{\small \textbf{Comparison of video-based person ReID methods}. Existing solutions typically process videos to extract the most overlapping and robust features, such as long-term features (\eg, static body shape), which may not adequately address all identification scenarios, especially when dynamic attributes like gait are absent. Our \textbf{HAMoBE} approach dynamically adjusts the processing of features based on the specific context of paired video inputs, effectively integrating long-term, short-term, and temporal features to enhance identification accuracy and adaptability.}
  \label{fig:teaser}
\end{figure}
%---------------------------------------------------------

Extracting robust features ($f$) across diverse variations poses a significant challenge in video-based person ReID. Accurately capturing and integrating three key features: long-term feature that is consistent over time (\emph{i.e.}, static body shape), short-term feature that changes over days (\emph{i.e.}, appearance), and temporal feature that varies dynamically within a video (\emph{i.e.}, gait) are essential to ensure reliable identification under varying conditions such as changes in lighting, background, camera views, clothing alterations, and human activities. 
Each of these discriminative features plays a dynamically significant role depending on the scenario: \emph{short-term feature} is particularly critical in short-term recognition where clothing and visual details are consistent, while \emph{temporal feature} becomes a key identifier in situations where individuals are moving, such as walking or running. \emph{Long-term feature} provides a consistent baseline that remains useful across various conditions, especially when other features may be obscured or altered. 
\emph{Adaptively} \emph{\textbf{prioritizing}} and \emph{\textbf{fusing}} these features based on video context are crucial for the effectiveness of ReID systems.

Existing video-based ReID methods predominantly utilize RNN/LSTM, 3D convolution, and attention mechanisms to aggregate temporal and spatial features~\cite{mclaughlin2016recurrent,dai2018video,li2019multi,gu2020appearance,liu2021watching}. 
While they are adept at generating robust features, they frequently fail to adapt dynamically to the shifting contexts presented in different videos. In particular, they often struggle to prioritize and integrate critical discriminative features effectively. 
Consequently, in scenarios where the temporal feature is absent and individuals appear stationary with varying clothing across videos, models trained with conventional settings may prioritize the extraction of the most overlapping and stable features, commonly $f_L$ (long-term feature). 
This choice can lead to the neglect of $f_S$ (short-term feature) or $f_T$ (temporal feature), each of which can be vital for identity recognition under diverse conditions (see Fig.~\ref{fig:teaser}).

To address these challenges, we propose a novel approach for video-based person ReID, termed Hierarchical and Adaptive Mixture of Biometric Experts (\textbf{HAMoBE}). As shown in Fig.~\ref{fig:teaser}, this framework leverages a Mixture of Experts (MoE) mechanism~\cite{jacobs1991adaptive} to dynamically and adaptively select and prioritize the three features—$f_L$, $f_S$, and $f_T$—based on the specific context presented by each pair of videos. 
By assessing the relevance and reliability of each feature between paired video inputs, HAMoBE ensures that the most critical features are accentuated, thereby enhancing the system's adaptability and significantly improving identification accuracy across diverse scenarios.

Specifically, the HAMoBE framework begins with a pretrained large-scale vision model (\emph{e.g.}, CLIP~\cite{radford2021learning}) that extracts multi-layer features, forming a rich base.
The key design lies in a two-level hierarchical mixture of biometric experts. The first level includes several experts (\eg, $8$) that process the raw features extracted by CLIP, distilling fundamental characteristics such as motion patterns, basic shape outlines, and general appearance cues.
These low-level features are then processed in the second level, where specialized experts combine them into high-level discriminative attributes corresponding to the long-term feature ($f_L$), the short-term feature ($f_S$), and the temporal feature ($f_T$).
To ensure robust matching, we devise a novel dual-input decision gating network that dynamically adjusts the contributions of each expert based on the relevance of the input scenarios, prioritizing and fusing the most pertinent features for effective matching. 
Experiments demonstrate the superiority of our method across diverse video-based person ReID benchmarks, achieving significant performance improvements.

In summary, the contributions of this work include:

$\diamond$ We propose \textbf{HAMoBE}, a novel adaptive approach for video-based person ReID that dynamically integrates long-term, short-term, and temporal features to enhance identification accuracy.

$\diamond$ We devise a novel two-level hierarchical mixture of biometric experts framework that effectively distills low-level features and disentangles them into high-level discriminative attributes.

$\diamond$ Extensive experiments demonstrate the superiority of \textbf{HAMoBE} in video-based person ReID.

%% file: sec/2_prior.tex
%---------------------------------------------------------
\begin{figure*}[t]
  \centering
  \resizebox{0.9\linewidth}{!}{
  \includegraphics[trim=0 0 0 0,clip]{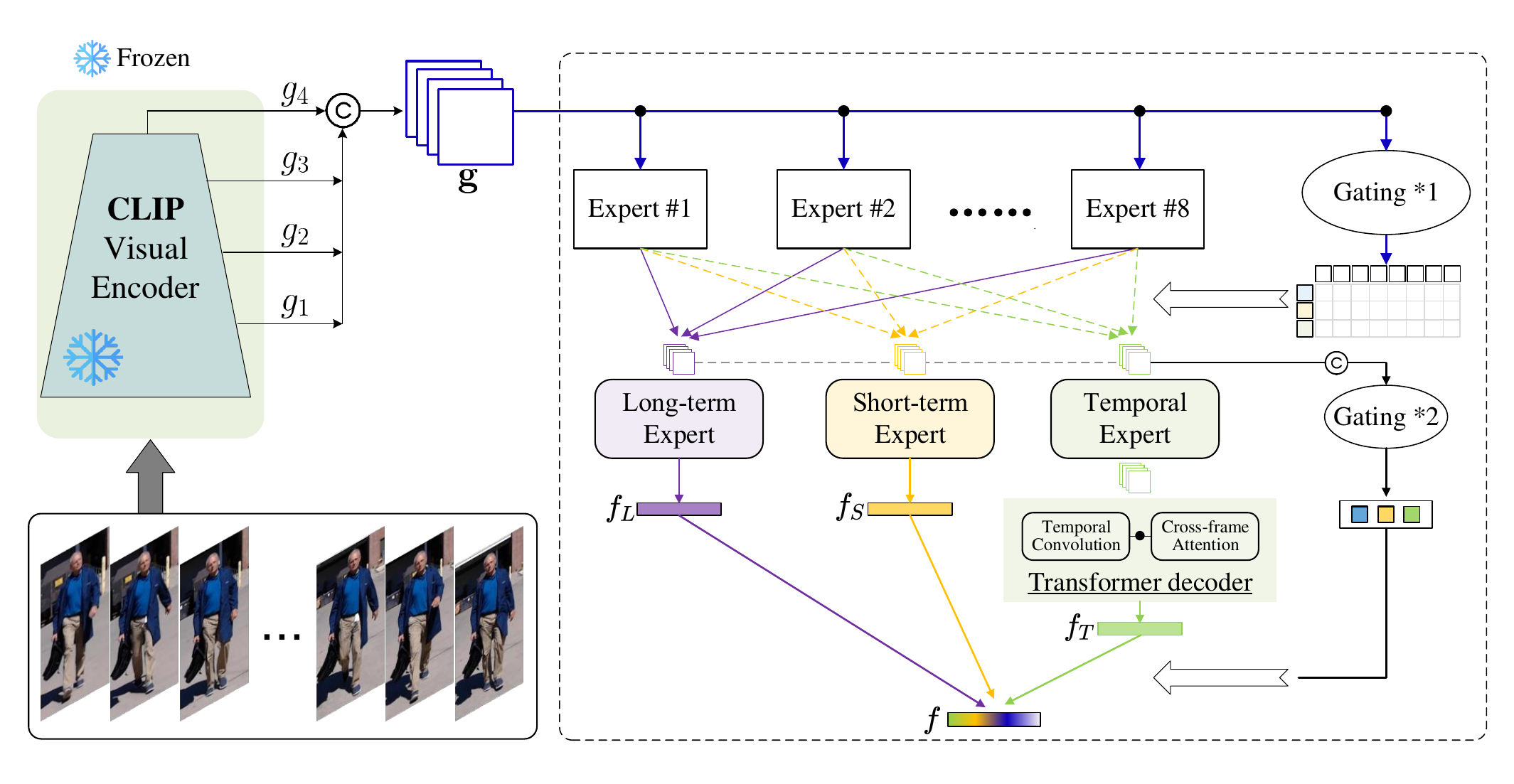}
  }
  \vspace{-5mm}
  \caption{\small \textbf{Overview of the proposed HAMoBE}. Starting with the frozen CLIP Visual Encoder, which processes input videos into multi-layer feature maps ($g_1, g_2, g_3, g_4$). These raw maps are then concatenated and passed through a hierarchical system of biometric experts. The first layer of experts processes raw features into more distinct features, which are then dynamically weighted by a gating network. This preparatory output ($\mathcal{F}_1, \mathcal{F}_2, \mathcal{F}_3$) feeds into the second-layer experts, who refine these signals into disentangled representations for long-term $f_L$, short-term $f_S$, and temporal $f_T$ features. The second-layer gating network integrates these into a final feature vector, $f$, used for person identification, dynamically adjusting the contributions based on video context.}
  \label{fig:overview}
\end{figure*}
%---------------------------------------------------------

\section{Related Work}
\Paragraph{Image-based Person ReID}
Image-based person ReID aims to identify a person across images captured by a distributed camera system.
Most existing approaches~\cite{roth2014exploration,ge2020mutual,ge2020self,li2018unsupervised,li2019unsupervised,lin2019bottom,wang2018transferable,yu2019unsupervised,zhai2020ad,ag-reid-2023-aerial-ground-person-re-identification-challenge-results} focus on short-term scenarios where clothing changes are not considered. This limitation has sparked interest in long-term cloth-changing person ReID~~\cite{gu2022clothes,yang2019person,li2021learning,jin2022cloth,su2024open,kim2025sapiensid}. 
Datasets such as Real28~\cite{wan2020person}, VC-Clothes~\cite{wan2020person}, PRCC~\cite{yang2019person}, LTCC~\cite{shu2021large}, COCAS~\cite{yu2020cocas}, Celebrities-reID~\cite{huang2019celebrities,huang2019beyond}, LaST~\cite{shu2021large} and DeepChange~\cite{xu2021deepchange} have been created to support this line of research. 
Building on these datasets, recent approaches~\cite{gu2022clothes,huang2019beyond,yu2020cocas,huang2019celebrities,yang2019person,shu2021large,li2021learning,jin2022cloth} have been proposed to investigate long-term person ReID under clothing changes. They extract clothing-irrelevant features for robust person ReID by custom-designed architectures~\cite{huang2019beyond,huang2021clothing}, training process~\cite{li2021learning}, loss functions~\cite{gu2022clothes}, data augmentation~\cite{zheng2019joint} or the incorporation of auxiliary 3D shape constraints~\cite{zheng2020parameter,chen2021learning,liu2023learning,liudistilling}.
Building on these foundational studies, our work extends these methodologies by applying a novel video-based framework that 
harnesses temporal dynamics.

\Paragraph{Video-based Person ReID}
Unlike image-based ReID, which relies on images, video-based ReID exploits both spatial and temporal information to handle the dynamic nature of video data.
Most existing methods~\cite{yu2024tf,nguyen2024temporal,kim2023feature} leverage temporal and spatial features via RNNs/LSTMs~\cite{mclaughlin2016recurrent,dai2018video}, 3D convolutions~\cite{li2019multi,gu2020appearance}, and attention mechanisms~\cite{liu2021watching}.
While these techniques capture not only spatial details but also temporal dynamics from video sequences, they struggle to effectively isolate and utilize the most discriminative biometric features adaptively, particularly in environments with frequent appearance changes or significant dynamics. 
To tackle these challenges, our HAMoBE innovatively separates and adaptively focuses on long-term, short-term, and temporal features and employs a dual-input decision gating network to dynamically fine-tune feature integration based on their relevance, improving the accuracy and reliability of video-based ReID.

\Paragraph{Gait Recognition}
Gait recognition, a method for identifying subjects in videos, relies on body shape and dynamic pose information. Mainstream approaches \cite{GaitSet_2019_AAAI,chao2021gaitset,GaitPart_2020_CVPR,GaitGL_2021_ICCV,GaitGCI_CVPR_2023,GaitMask_2021_BMVC,MetaGait,OpenGait_CVPR_2023,DyGait_ICCV_2023,LidarGait_CVPR_2023,LagrangeGait_2022_CVPR,3DLocalGait_2021_ICCV,GLN_2020_ECCV,kim2024keypoint} typically use silhouettes as input to filter out irrelevant details such as clothing and texture. However, silhouettes provide limited information. Recently, some gait recognition methods have directly utilized gait information from RGB images \cite{GaitEdge,GaitNet,zhang2019gait,ye2024biggait}. These methods exploit the rich information in RGB images, capturing not only the silhouette but also color and texture details, thereby enhancing the model's discriminative power. In video-based person ReID, gait information serves as a crucial source of invariant information, unaffected by variables like clothing.

\Paragraph{Mixture of Experts (MoE)}
In machine learning, MoE stands as a powerful paradigm for addressing complex tasks by blending the expertise of multiple models. Originating from the seminal work of~\cite{jacobs1991adaptive, jordan1994hierarchical}, MoE architectures have garnered significant attention for their performance and interpretability. By organizing a network into specialized subnetworks, or ``experts,'' each responsible for different regions of input space, MoE efficiently captures diverse patterns within data. This approach has found application across various domains, including natural language processing~\cite{jiang2024mixtral, fedus2022switch, zoph2022st, komatsuzaki2022sparse, li2024merge, shazeer2017outrageously, lepikhin2020gshard}, computer vision~\cite{chen2023adamv, eigen2013learning, ahmed2016network, gross2017hard, wang2020deep, yang2019condconv, abbas2020biased, pavlitskaya2020using, riquelme2021scaling, chen2024eve, you2024shiftaddvit}, and biometrics~\cite{wang2025decoupled, wang2021multi, dai2021generalizable,jawade2024proxyfusion, zhu2025quality}.
In biometrics, one alternative to MoE is multi-modality fusion~\cite{improving-biometric-identification-through-quality-based-face-and-fingerprint-biometric-fusion,improving-face-recognition-with-a-quality-based-probabilistic-framework}.
Drawing inspiration from this idea, we apply the MoE concept to video-based person ReID, focusing on the dynamic and adaptive assignment of feature contributions. 
Our adaptation allows for precise and adaptable recognition by dynamically adjusting how each feature's contribution is weighed based on the context of the input video pair. This approach ensures more context-aware and accurate person ReID.

%% file: sec/3_method.tex
\section{Methodology}

As shown in Fig.~\ref{fig:overview}, HAMoBE consists of two main parts: the upstream frozen large vision model (\emph{e.g.}, CLIP) for raw multi-layer feature extraction and the two-level hierarchical mixture of biometric experts responsible for disentangling high-level discriminative features.
HAMoBE takes an RGB video or paired RGB videos as input and processes each frame in parallel. Before feeding the images into the pre-trained large model, a Pad-and-Resize technique is applied to resize them to a fixed resolution of $224\times224$ while maintaining the original body's aspect ratio.

\subsection{Multi-Layer Raw Feature Extraction}

We utilize the frozen CLIP visual encoder~\cite{radford2021learning} to extract multi-layer features from a resized RGB video $\mathbf{V}\in\mathbb{R}^{T\times C\times H\times W}$ ($T$ denotes the number of frames, $C$ is the color channel), given its robust performance in diverse visual tasks.
For each frame $\mathbf{V}_{t}$, the Vision Transformer (ViT) within CLIP visual encoder first segments it into non-overlapping patches of $14\times14$ pixels, producing $16\times16=256$ tokenized vectors.
As illustrated in Fig.~\ref{fig:overview}, feature maps $\mathbf{G}_1$, $\mathbf{G}_2$, $\mathbf{G}_3$, $\mathbf{G}_4\in\mathbb{R}^{T\times 257\times d}$, where \(T\) is the number of frames, $d$ is the embedding dimension of the tokens, and \(257\) corresponds to \(256\) image tokens and the \verb|CLS| token, are generated at different layers of the ViT backbone to capture a hierarchy of semantic levels. We concatenate them to form a feature volume $\mathbf{G}=\operatorname{concat}(\mathbf{G}_1, \mathbf{G}_2, \mathbf{G}_3, \mathbf{G}_4)\in\mathbb{R}^{T\times 257\times 4d}$, following~\cite{lin2022frozen}.

\subsection{Hierarchical Mixture of Biometric Experts}

Our proposed Hierarchical Mixture of Biometric Experts network consists of two layers, each equipped with a distinct set of experts and a gating network to dynamically value their contributions.

\Paragraph{First-Layer Experts}
The first layer comprises $n_1$ biometric experts ($n_1=8$ in our experiments), implemented as Multi-Layer Perceptrons (MLPs). These experts extract specific low-level features from the raw features extracted by CLIP, distilling fundamental characteristics such as motion patterns, basic shape outlines, and general appearance cues. Each expert $\mathcal{E}^{1}_{i}$ processes these extracted multi-layer raw features $\mathbf{G}$, resulting in a series of outputs:
\begin{equation}
    \mathbf{F}_i = \mathcal{E}^{1}_{i} (\mathbf{G}), i=1, \dots, n_1, \label{eqn:first_layer}
\end{equation}
where $\mathbf{F}_i\in\mathbb{R}^{T\times 257 \times d}$ is the output feature volume from the corresponding expert $\mathcal{E}^{1}_{i}$. These outputs are then dynamically weighted by the first-layer gating network. 

\Paragraph{First-Layer Gating Network $\Phi^{*1}$} 
Concurrently, this gating network $\Phi^{*1}$ adjusts the contributions of the first-layer experts, producing a tensor $\mathbf{W}^{*1}=\Phi^{*1}(\mathbf{G})\in\mathbb{R}^{T\times 257 \times n_1\times n_2}$, where $n_2$ denotes the number of experts in the second layer ($n_2=3$ in our experiments). The output is a weighted combination of the first-layer expert outputs, forming the input for the second layer:
\begin{equation}
    \mathbf{F}^{(j)} = \sum_{i=1}^{n_1} \left( \mathbf{W}^{*1}_{i,j} \odot \mathbf{F}_i \right), \quad j = 1, \dots, n_2,
\end{equation}
where $\mathbf{W}^{*1}_{i,j} = \mathbf{W}^{*1}[:,:,i,j]$ denotes the weights, and $\mathbf{F}^{(j)}$ is the input feature volume for the \(j\)-th second-layer experts. ``$\odot$'' denotes element-wise multiplication (\(\mathbf{W}^{*1}_{i,j}\) is broadcast in the last/channel dimension to match \(\mathbf{F}^{(j)}\)).

\Paragraph{Second-Layer Experts}
The second layer consists of specialized expert modules, each tasked with distilling low-level features and disentangling them into high-level discriminative attributes:

\quad$\bullet$ \emph{Long-Term Expert} $\mathcal{E}_{L}$: Focuses on the invariant aspects of the human body shape and other enduring characteristics. The model output 
\(
\mathbf{F}_L = \mathcal{E}_L(\mathbf{F}^{(1)}) \in \mathbb{R}^{T \times d}
\)
is obtained through an MLP, which aggregates information across the patch dimension. Subsequently, an average pooling operator $\textup{AP}(\cdot)$ is applied to yield
\(
\mathbf{f}_L = \textup{AP}(\mathbf{F}_L) \in \mathbb{R}^{d},
\)
averaging across the frame dimension.

\quad$\bullet$ \emph{Short-Term Expert} $\mathcal{E}_{S}$: Targets detailed visual attributes such as clothing texture, color, and other per-video distinguishing cues. The output 
\(
\mathbf{F}_S = \mathcal{E}_S(\mathbf{F}^{(2)}) \in \mathbb{R}^{T \times d}
\)
is produced by an MLP that aggregates the patch dimension. An $\textup{AP}(\cdot)$ is then employed to derive
\(
\mathbf{f}_S = \textup{AP}(\mathbf{F}_S) \in \mathbb{R}^{d}.
\)

\quad$\bullet$ \emph{Temporal Expert} $\mathcal{E}_{T}$: Analyzes poses and captures dynamic motion over time. The feature volume,
\(
\mathbf{F}_T = \mathcal{E}_T(\mathbf{F}^{(3)}) \in \mathbb{R}^{T \times d},
\)
is computed by an MLP, followed by a Transformer decoder. The operations of the Transformer decoder are formally expressed as~\cite{lin2022frozen}:
\begin{align}
\mathbf{Y}_i &= \operatorname{Temp}_i([\mathbf{f}_{T}^{1}, \dots, \mathbf{f}_{T}^{t}]), \notag \\
\tilde{\mathbf{q}}_i &= \mathbf{q}_{i-1} + \operatorname{MHSA}_i(\mathbf{q}_{i-1}, \mathbf{Y}_i, \mathbf{Y}_i), \notag \\
\mathbf{q}_i &= \tilde{\mathbf{q}}_i + \operatorname{MLP}_i(\tilde{\mathbf{q}}_i), \notag \\
\mathbf{f}_T &= \operatorname{FC}(\mathbf{q}_M),
\label{eqn:pose}
\end{align}
where \(\mathbf{f}_{T}^{t}\) represents the temporal (\eg, gait) features of frame \(t\). The feature volume \(\mathbf{Y}_i\), which undergoes temporal modulation, is fed into the \(i\)-th layer of the Transformer decoder. The query token \(\mathbf{q}_i\) is incrementally refined, beginning with \(\mathbf{q}_0\) as learnable initial parameters. The final output 
\(
\mathbf{f}_{T}\in\mathbb{R}^d
\) corresponds to the gait feature. The decoder comprises $M$ blocks. The multi-head self-attention (MHSA) module models the relations among the tokens. The operator $\operatorname{Temp}(\cdot)$ models temporal dynamics and produces feature tokens with detailed temporal information.

\Paragraph{Second-Layer Gating Network $\Phi^{*2}$}  This gating network component is critical for synthesizing the comprehensive identification profile by dynamically aggregating and fusing the key features—long-term $f_{L}$, short-term $f_{S}$, and temporal $f_{T}$—extracted by the second-layer experts. It generates three distinct weights: $[{w}^{*2}_L, {w}^{*2}_S, {w}^{*2}_T] = \Phi^{*2}(\text{concat}(\mathbf{F}^{(1)}, \mathbf{F}^{(2)}, \mathbf{F}^{(3)}))$, allowing for an optimized combination of these features based on the current identification scenario. 
The $\Phi^{*2}$ is implemented using an MLP followed by a softmax layer, which normalizes the output weights to ensure they sum to $1$.
Formally, the operational formula for this dynamic fusion process can be expressed as follows:
\begin{equation}
    \mathbf{f} = {w}^{*2}_L \cdot \mathbf{f}_L + {w}^{*2}_S \cdot \mathbf{f}_S + {w}^{*2}_T \cdot \mathbf{f}_T.
    \label{eqn:second_layer_weight}
\end{equation}
This adaptive weighting mechanism ensures that the contributions of each feature are precisely tailored to the specific requirements of context-aware person identification.
%---------------------------------------------------------------------------

%---------------------------------------------------------------------------
\begin{figure}[t]
\centering
\begin{subfigure}[b]{0.45\textwidth}
   \includegraphics[width=\linewidth]{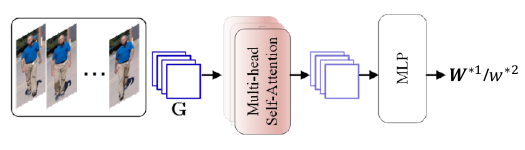}
   \caption{Single-input mode.}
   \label{fig:gating_a}
\end{subfigure}
\begin{subfigure}[b]{0.45\textwidth}
   \includegraphics[width=\linewidth]{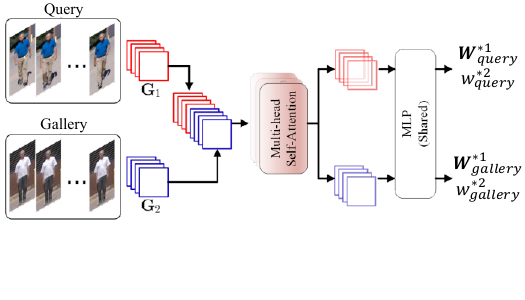}
   \vspace{-1.4cm}
   \caption{Dual-input mode.}
   \label{fig:gating_b}
\end{subfigure}
\caption{\small\textbf{The proposed dual-input gating network in HAMoBE}. This setup enables the network to adaptively synthesize information from both inputs, enhancing identification accuracy by leveraging contextual differences between the paired videos. Note that Fig.~\ref{fig:overview} illustrates the single-input mode. In the dual-input mode, both inputs are processed in parallel by all modules except the two gating networks.
}
\label{fig:subfigures}
\end{figure}
%---------------------------------------------------------------------------

\subsection{Dual-Input Decision Mechanism}

\Paragraph{Dual-Input Gating Network} 
We devise a dual-input gating network to handle both single-input (Fig.~\ref{fig:gating_a}) and dual-input scenarios, adapting its processing strategy based on the input. This flexibility allows the network to process and integrate features under varying contextual conditions optimally.
As illustrated in Fig.~\ref{fig:gating_b}, when two video sequences $\mathbf{V}_{\textup{gallery}}$ and $\mathbf{V}_{\textup{query}}$ are available, the network first concatenates their multi-layer raw feature along the spatial dimension $[\mathbf{G}_{\textup{gallery}}, \mathbf{G}_{\textup{query}}]=\operatorname{concat}(\mathbf{G}_{\textup{gallery}}, \mathbf{G}_{\textup{query}})\in\mathbb{R}^{T\times 257\times {8}d}$. The concatenated features are then processed by MHSA, which extracts relevant features from the input:
\begin{equation}
    [\mathbf{G}_\textup{gallery}', \mathbf{G}_\textup{query}'] = \operatorname{MHSA}\left([\mathbf{G}_\textup{gallery}, \mathbf{G}_\textup{query}]\right). 
    \label{eqn:dual_input}
\end{equation}
Separate MLPs subsequently process the output of MHSA to determine the weights for integrating these features:
\begin{align}
    {w}_{\textup{gallery}} = \operatorname{MLP}(\mathbf{G}_{\textup{gallery}}'), \quad {w}_{\textup{query}} = \operatorname{MLP}(\mathbf{G}_{\textup{query}}').
\end{align}
These weights are then applied to adjust the contributions from each input, synthesizing a feature vector that embodies the most informative aspects of both videos. This method ensures that the system can make robust decisions based on the paired features, enhancing the identification accuracy by dynamically leveraging the context of paired video inputs.

\emph{It is noteworthy that this dual-input gating mechanism is seamlessly applicable to both layers of the gating network, $\Phi^{*1}$ and $\Phi^{*2}$. During training, both single-input mode and dual-input mode are utilized, enhancing the model’s adaptability and robustness across diverse scenarios.} 

\Paragraph{Dual-Input Decision Mechanism in Inference} 
This Dual-Input Decision Mechanism can also be applied for evaluation, leveraging weights $\mathcal{W}_{\textup{gallery}}$ and $\mathcal{W}_{\textup{query}}$ from the MHSA module to integrate features from the dual video inputs $\left(\mathbf{V}_{\textup{gallery}}, \mathbf{V}_{\textup{query}}\right)$. The integrated features form composite vectors for each sequence, which are then compared against known identity profiles using a similarity metric, such as cosine similarity.
Applying the dual-input decision mechanism to the central \( q\% \) of query-gallery pairs—ranging from the \( \left(50 - \frac{q}{2}\right)^{\text{th}} \) to the \( \left(50 + \frac{q}{2}\right)^{\text{th}} \) percentile—maximizes the adaptive capabilities of the dual-input gating network, ensuring robust performance in complex and variable environments.
This not only enhances the precision of identity matches but also refines the system’s ability to adapt to a range of operational contexts.

\subsection{Model Learning}
The overall training loss function is as follows:
\begin{multline}
    \mathcal{L} = \mathcal{L}_{CE} + \alpha(\mathcal{L}_\textup{LTS} + \mathcal{L}_\textup{STS} + \mathcal{L}_\textup{TS}) + \beta\mathcal{L}_{\text{contrastive}},
    \label{eqn:objective}
\end{multline}
where $\mathcal{L}_{CE}$ is the cross-entropy loss on feature vector $f$, $\mathcal{L}_\textup{LTS}$ is long-term consistency loss, $\mathcal{L}_\textup{STS}$ denotes short-term consistency loss, $\mathcal{L}_\textup{TS}$ is the temporal loss, $\mathcal{L}_{\text{contrastive}}$ is the contrastive loss, and $\alpha$ and $\beta$ are weights assigned to balance the loss terms. 

\Paragraph{Long-term Consistency Loss} The static body shape feature $f_{B}$ describes the subject's consistent body characteristics across all video frames. For two videos of the sample subjects, $\mathbf{V}_{1}$ and $\mathbf{V}_{2}$, their long-term feature (\emph{i.e.}, \emph{global} static body shape) $\mathbf{f}^1_L$ and $\mathbf{f}^2_L$ should be consistent. Also, the frame-based features $\mathbf{F}^{1 (t)}_L$ from $t=1$ to $T$ for $\mathbf{V}_1$ and $\mathbf{F}^{2 (t)}_L$ from $t=1$ to $T$ for $\mathbf{V}_2$, should each maintain consistency across their respective sequences.
{
\begin{multline}
\mathcal{L}_{\textup{LTS}} = \|\mathbf{f}_L^1 - \mathbf{f}_L^2\|_2^2 \\
+ \frac{1}{t^2} \sum_{t_1 \ne t_2} \left( \| \mathbf{F}_L^{1(t_1)} - \mathbf{F}_L^{1(t_2)} \|_2^2 + \| \mathbf{F}_L^{2(t_1)} - \mathbf{F}_L^{2(t_2)} \|_2^2 \right).
\label{eqn:static_sim}
\end{multline}
}
\Paragraph{Short-term Consistency Loss}
The short-term features (\emph{i.e.}, appearance), \( \mathbf{F}^{1(t)}_S \) and \( \mathbf{F}^{2(t)}_S \), from videos \( \mathbf{V}_1 \) and \( \mathbf{V}_2 \) respectively, should be consistent within each video sequence. We quantify this through the appearance consistency loss, which penalizes variance among these features across frames:
\begin{equation}
\resizebox{\columnwidth}{!}{
$\mathcal{L}_{\textup{STS}} = \frac{1}{t^2} \sum_{t_1 \ne t_2} \left( \| \mathbf{F}_S^{1(t_1)} - \mathbf{F}_S^{1(t_2)} \|_2^2 + \| \mathbf{F}_S^{2(t_1)} - \mathbf{F}_S^{2(t_2)} \|_2^2 \right)$.
}
\label{eqn:appe_sim}
\end{equation}

\Paragraph{Temporal Consistency Loss} 
The temporal consistency loss ensures the temporal coherence of gait features across video frames, which is critical for robust identification in dynamic conditions. It is formulated as
\begin{equation}
\mathcal{L}_{\textup{TS}} = \| \mathbf{f}_T^1 - \mathbf{f}_T^2 \|_2^2,
\label{eqn:temp_sim}
\end{equation}
where $\mathbf{f}^1_T$ and $\mathbf{f}^2_T$ are the gait features from two inputs. 

\Paragraph{Contrastive Loss for Dual-Input Mode} 
In dual-input mode training, a contrastive loss is applied to the final feature vector $f$, enhancing the model's capability to distinguish between different individuals while affirming similarity when comparing the same individual across different sequences. This loss is formulated as:
\begin{equation}
\resizebox{\columnwidth}{!}{
$\mathcal{L}_{\text{contrastive}} = \frac{1}{2} \left( y \left\lVert \mathbf{f}^1 - \mathbf{f}^2 \right\rVert_2^2 + (1 - y) \cdot \max\left(0, m - \left\lVert \mathbf{f}^1 - \mathbf{f}^2 \right\rVert_2 \right)^2 \right)$,
}
\end{equation}
where $\mathbf{f}^1$ and $\mathbf{f}^2$ are the final features from two inputs, $y$ is a binary label indicating if the inputs are from the same identity (1) or different identities (0), and $m$ is the margin parameter. This mechanism ensures robustness in identity verification across varied inputs.

\Paragraph{Implementation Details}
We use the ViT-L/14 model from CLIP as our multi-layer raw feature extractor, implemented in PyTorch with the Adam optimizer. Key parameters include $T=16$,  $\alpha=0.5$, $\beta=1$, $n_1=8$, $n_2=3$, $M=4$. 
Additionally, we set margin $m=4$ in $\mathcal{L}_{\text{contrastive}}$.

%% file: sec/4_exp.tex
\newcommand{\firstkey}[1]{\textcolor{RubineRed}{\textbf{#1}}}
\newcommand{\secondkey}[1]{\textcolor{NavyBlue}{\textbf{#1}}}

\begin{table}[t!]
\centering \small
\begin{tabular}{@{}lcccc@{}}
\toprule
\multirow{2}{*}{\textbf{Method}}  & \multicolumn{2}{c}{\textbf{MARS}} & \multicolumn{2}{c}{\textbf{LS-VID}} \\
\cmidrule(lr){2-3} \cmidrule(lr){4-5} 
                  &       mAP & top-1 &  mAP & top-1   \\
\midrule
PSTA~\cite{wang2021pyramid} & $85.8$ & $91.5$ & - & - \\
DIL~\cite{he2021dense} & $87.0$ & $90.8$ & - & - \\
STT~\cite{zhang2021spatiotemporal}  & $86.3$ & $88.7$ & $78.0$ & $87.5$  \\
TMT~\cite{liu2024video}  & $85.8$ & $91.2$ & $-$ & $-$ \\
CAVIT~\cite{wu2022cavit}  & $87.2$ & $90.8$ & $79.2$ & $89.2$  \\
SINet~\cite{bai2022salient} & $86.2$ & $91.0$ & $79.6$ & $87.4$ \\
MFA~\cite{gu2022motion} & $85.0$ & $90.4$ & $78.9$ & $88.2$ \\
DCCT~\cite{liu2023deeply} & $87.5$ & $92.3$ & - & - \\
LSTRL~\cite{liu2023video} & $86.8$ & $91.6$ & $82.4$ & $89.8$ \\
TF-CLIP~\cite{yu2024tf} & \secondkey{$89.4$} & \secondkey{$93.0$} & \secondkey{$83.8$} & \secondkey{$90.4$}
\\ \midrule
\textbf{HAMoBE}         & \firstkey{$91.1$} & \firstkey{$94.6$} & \firstkey{$85.2$} & \firstkey{$92.1$}  \\
\bottomrule
\end{tabular}
\caption{\small Performance comparison ($\%$) on the MARS~\cite{zheng2016mars} and LS-VID~\cite{li2019global} datasets. [Key: \firstkey{Best}, \secondkey{Second Best}] }
\label{tab:general_db}
\end{table}

\section{Experiments}

In this section, we evaluate our model's performance with a primary focus on video-based person re-identification. Additionally, we conduct ablation studies to analyze the contributions of individual components and visualize the effectiveness of our approach.

\subsection{Video-based Person ReID}

\Paragraph{Datasets} We evaluate on four widely-used video-based person ReID datasets: MARS~\cite{zheng2016mars}, LS-VID~\cite{li2019global}, CCVID~\cite{gu2022clothes}, and MEVID~\cite{davila2023mevid}. While MARS and LS-VID only have same-clothing (SC) protocols, CCVID and MEVID additionally provide protocols for more challenging \emph{different-clothing} (DC) settings.

\Paragraph{Baseline} Our \textbf{HAMoBE}, is rigorously evaluated against $26$ state-of-the-art (SoTA) methods. 
These include pyramid spatial transformer networks like PSTA~\cite{wang2021pyramid}, densely interactive networks like DIL~\cite{he2021dense}, spatiotemporal transformers such as STT~\cite{zhang2021spatiotemporal}, and advanced models like TMT~\cite{liu2024video}, CAVIT~\cite{wu2022cavit}, and SINet~\cite{bai2022salient}. Specialized methods for motion focus, such as MFA~\cite{gu2022motion} and deeply cascaded contextual trackers like DCCT~\cite{liu2023deeply}, are also included. Recent innovations like TF-CLIP~\cite{yu2024tf} (AAAI'24) and CLIP3DReID~\cite{liudistilling} (CVPR'24), which utilize the pre-trained CLIP model, highlight the growing trend of integrating large-scale pretrained models in ReID tasks. Additionally, gait recognition methods such as GaitNet~\cite{GaitNet} and GaitSet~\cite{GaitSet_2019_AAAI} are considered to provide a comprehensive analysis across different approaches and technologies in the field of person ReID.

\Paragraph{Metrics} We use standard retrieval accuracy metrics for evaluation, specifically Cumulative Matching Characteristics (CMC) and mean Average Precision (mAP).

\begin{table}[t]
\centering \small
\resizebox{\linewidth}{!}{
\begin{tabular}{@{}lcccc@{}}
\toprule
\multirow{2}{*}{\centering \textbf{Method}} & \multicolumn{2}{c}{\textbf{General (\%)}} & \multicolumn{2}{c}{\textbf{Different Clothes (\%)}} \\ 
\cmidrule(lr){2-3} \cmidrule(lr){4-5}
 & mAP & top-1 & mAP & top-1  \\  
\midrule
TriNet~\cite{hermans2017defense}   & $78.1$  & $81.5$ & $77.0$ & $81.1$ \\
I3D~\cite{carreira2017quo}   & $79.7$ & $76.9$ & $78.5$ & $75.3$ \\
Non-Local~\cite{wang2018non}   & $80.7$ & $78.0$ & $79.3$ & $76.2$ \\
TCLNet~\cite{hou2020temporal}   & $81.3$ & $77.9$ & $80.7$ & $75.9$ \\
AP3D~\cite{gu2020appearance}   & $80.9$ & $79.2$ & $80.1$ & $77.7$ \\
CAL~\cite{gu2022clothes}   & $81.3$ & $82.6$ & $79.6$ & $81.7$ \\
3DInvarReID~\cite{liu2023learning}    & $82.6$ & $83.9$ & $81.3$ & $81.7$ \\
CLIP3DReID~\cite{liudistilling}    & \secondkey{$83.9$} & \secondkey{$84.5$} & \secondkey{$83.2$} & \secondkey{$82.4$} 
\\ \midrule
GaitNet~\cite{GaitNet}    & $62.6$ & $56.5$ & $57.7$ & $49.0$ \\
GaitSet~\cite{GaitSet_2019_AAAI}    & $81.9$ & $73.2$ & $71.2$ & $62.1$ 
\\ \midrule
\textbf{HAMoBE}   & \firstkey{$87.2$} & \firstkey{$88.4$} & \firstkey{$85.8$} & \firstkey{$87.6$} \\
\bottomrule
\end{tabular}
}
\caption{\small Performance comparison on the CCVID dataset~\cite{gu2022clothes}. [Key: \firstkey{Best}, \secondkey{Second Best}] }
\label{tab:ccvid}
\vspace{-2mm}
\end{table}

\begin{table*}[t]
\centering

\resizebox{0.8\linewidth}{!}{
\begin{tabular}{@{}lccccccccccc@{}}
\toprule
\multirow{2}{*}{\centering \textbf{Method}} & \multicolumn{5}{c}{\textbf{Same Clothes (\%)}} & \multicolumn{5}{c}{\textbf{Different Clothes (\%)}} \\ 
\cmidrule(lr){2-6} \cmidrule(lr){7-11}
  & mAP & top-1 & top-5 & top-10 & top-20 & mAP & top-1 & top-5 & top-10 & top-20 \\ 
\midrule
AGRL~\cite{wu2020adaptive}        & $32.6$ & $51.4$ & $64.9$ & $73.6$ & $80.9$ & {$5.7$} & $4.9$ & {$15.1$} & $19.0$ & $25.7$ \\
BiCnet-TKS~\cite{hou2021bicnet}  &  $8.0$ & $20.5$ & $36.5$ & $41.7$ & $51.4$ & $1.7$ & $0.7$ &  $4.6$ &  $7.8$ & $13.4$ \\
TCLNet~\cite{hou2020temporal}      & $31.9$ & $51.7$ & $63.5$ & $71.9$ & $79.2$ & $3.9$ & $3.5$ &  $8.8$ & $14.1$ & $21.1$ \\
PSTA~\cite{wang2021pyramid}        & $29.7$ & $49.0$ & $63.9$ & $72.2$ & $78.5$ & $5.1$ & {$5.6$} & $12.3$ & {$19.4$} & {$28.9$} \\
PiT~\cite{zang2022multidirection}         & $19.5$ & $36.8$ & $58.7$ & $66.3$ & $73.6$ & $2.0$ & $1.1$ &  $5.3$ &  $8.5$ & $13.7$ \\
STMN~\cite{eom2021video}        & $18.5$ & $33.7$ & $58.3$ & $69.1$ & $76.4$ & $1.2$ & $0.4$ &  $1.8$ &  $3.9$ &  $6.0$ \\
Attn-CL~\cite{pathak2020video}     & $24.2$ & $44.4$ & $59.7$ & $66.3$ & $72.6$ & $3.4$ & $2.8$ &  $8.5$ & $15.5$ & $24.6$ \\
Attn-CL+rerank~\cite{pathak2020video} & $34.1$ & $50.7$ & $63.2$ & $68.1$ & $72.9$ & $4.2$ & $2.1$ &  $9.2$ & $13.7$ & $22.5$ \\
AP3D~\cite{gu2020appearance}        & $23.2$ & $42.7$ & $59.7$ & $67.7$ & $79.2$ & $2.9$ & $1.8$ &  $7.4$ &  $9.5$ & $16.6$ \\
CAL~\cite{gu2022clothes}         & $39.0$ & {$56.6$}& {$70.8$} & {$78.1$} & {$85.4$} & $4.3$ & $3.5$ & $10.6$ & $14.8$ & $19.4$ \\
CLIP3DReID~\cite{liudistilling}         & \secondkey{$39.6$} & \secondkey{$60.9$}& \secondkey{$72.3$} & \secondkey{$79.4$} & \secondkey{$85.9$} & \secondkey{$9.6$}  & \secondkey{$11.2$} & \secondkey{$19.4$} & \secondkey{$25.3$} & \secondkey{$34.1$} \\\midrule
\textbf{HAMoBE}           & \firstkey{$54.3$} & \firstkey{$74.0$} & \firstkey{$84.7$} & \firstkey{$86.1$} & \firstkey{$88.2$} & \firstkey{$15.1$} & \firstkey{$20.1$} & \firstkey{$29.2$} & \firstkey{$38.4$} & \firstkey{$48.9$} \\
\bottomrule
\end{tabular}
}
\caption{\small Performance comparison on the MEVID dataset~\cite{davila2023mevid}. [Key: \firstkey{Best}, \secondkey{Second Best}]
\vspace{-2mm}}
\label{tab:mevid}
\end{table*}

\subsection{Results}

\Paragraph{MARS and LS-VID Datasets}
Tab.~\ref{tab:general_db} presents the performance of our HAMoBE model in comparison to state-of-the-art baselines on the MARS and LS-VID datasets. On {MARS}, HAMoBE surpasses the SoTA TF-CLIP model~\cite{yu2024tf}, by {1.7\%} on mAP and {1.6\%} on top-1 accuracy. On {LS-VID}, HAMoBE outperforms TF-CLIP by {1.4\%} in mAP and {1.7\%} in top-1.

The improvements are significant, particularly given that MARS is relatively saturated with limited variation in clothing and scene context. These results indicate that HAMoBE maintains strong accuracy under controlled conditions.
In contrast, LS-VID has more changes in lighting, background clutter, and occlusions. The slightly larger gains on LS-VID suggest that HAMoBE benefits from its hierarchical and adaptive MoE framework, which can dynamically emphasize informative regions, making it especially effective in unconstrained real-world video scenarios.

\Paragraph{CCVID Dataset}
As shown in Tab.~\ref{tab:ccvid}, HAMoBE achieves state-of-the-art performance on the {CCVID} dataset. Under the {SC protocol}, HAMoBE surpasses all previous methods, including the prior best CLIP3DReID. Under the more challenging {DC protocol}, HAMoBE achieves {85.8\%} mAP and {87.6\%} top-1 accuracy.

The performance drop from the SC to the DC protocol is smaller for HAMoBE than the baselines, highlighting its clothing invariance. This suggests that HAMoBE's expert routing promotes feature specialization of experts independent of clothing changes. 
Moreover, the relative performance gains over prior methods are larger than on MARS and LS-VID, underscoring HAMoBE’s particular strength under variable attire.

\Paragraph{MEVID Dataset}
Tab.~\ref{tab:mevid} details HAMoBE's performance on the {MEVID} dataset, a challenging benchmark due to significant variations in clothing, environments, and camera viewpoints. In the {SC} setting, HAMoBE achieves improvements of {14.7\%} in mAP and {13.1\%} in top-1 over the SoTA CAL model. In the more difficult {DC} scenario, HAMoBE maintains its lead with an mAP of {15.1\%} and top-1 accuracy of {20.1\%}, outperforming the second-best method by {5.5\%} and {8.9\%}, respectively.

The performance gain is pronounced in the SC setting. This can be attributed to HAMoBE’s ability to exploit visual consistency through fine-grained feature modeling. However, in the DC scenario, HAMoBE’s adaptive gating and hierarchical routing still outperform the baselines, but the relative margin is narrower due to the inherent challenge.

Among all datasets, MEVID shows the {largest absolute performance gains}, which better highlight the advantages of HAMoBE’s architecture. Competing methods struggle under such variability, while HAMoBE's MoE framework dynamically integrates different biometric experts and yields robust generalization in unconstrained conditions.

\subsection{Analysis and Visualization}
\Paragraph{Performance of Each Expert}
Tab.~\ref{tab:expert_ablation} shows the performance of each expert within our HAMoBE framework on the CCVID dataset. 
In our hierarchical MoE, first-layer experts provide a diverse basis of low-level features, 
supporting flexible composition by higher-level biometric experts. Although not directly supervised, Fig.~\ref{tab:expert_ablation} shows
their contributions to higher-level features are {\emph{complementary with minimal overlap}},
suggesting functional alignment with long-term, short-term, and temporal cues.

Of the second level experts, the temporal expert demonstrates the highest effectiveness, achieving over $85\%$ (top-1 accuracy) across all protocols, benefiting from the dataset’s focus on gait recognition, which provides rich temporal insights. The long-term expert shows consistent performance, while the short-term expert lags, particularly under the DC protocol, indicating its sensitivity to changes in clothing. By effectively combining the outputs of all experts, HAMoBE enhances overall performance, outstripping the capabilities of any single expert.

\begin{table}[t]
\centering \small

\begin{tabular}{@{}lcccc@{}}
\toprule
\multirow{2}{*}{\centering \textbf{Expert}} & \multicolumn{2}{c}{{General}} & \multicolumn{2}{c}{{Different Clothes}} \\ \cmidrule(lr){2-3} \cmidrule(lr){4-5}
 & mAP & top-1 & mAP & top-1  \\ \midrule
Long-term & \(57.5\) & \(64.7\) & \(50.3\) & \(56.0\) \\
Short-term & \(51.0\) & \(54.3\) & \(38.5\) & \(38.7\) \\
Temporal & {\(86.4\)} & {\(86.5\)} & {\(84.6\)} & {\(85.4\)} \\ \midrule
Single-Input Only & \(85.9\) & \(86.0\) & \(85.1\) & \(85.7\) \\ \midrule
\textbf{HAMoBE}   & \firstkey{$87.2$} & \firstkey{$88.4$} & \firstkey{$85.8$} & \firstkey{$87.6$} \\
\bottomrule
\end{tabular}
\caption{\small Analysis on different experts and dual-input decision mechanism on CCVID~\cite{gu2022clothes}.}
\label{tab:expert_ablation}
\end{table}
\begin{table}[t]
\centering \small
\begin{tabular}{@{}lcccccc@{}}
\toprule
\multirow{2}{*}{\centering {MoE}} & \multirow{2}{*}{\centering {Con.}} & \multicolumn{2}{c}{{Same Clothes}} & \multicolumn{2}{c}{{Different Clothes }} \\ 
 \cmidrule(lr){3-4} \cmidrule(lr){5-6}
  & Loss & mAP & top-1 & mAP & top-1 \\ 
\midrule
Vanilla & \ding{55} & \(36.6\) & \(63.0\) & \(10.6\) & \(14.3\) \\
HAMoBE & \ding{55} & \secondkey{\(50.4\)} & \secondkey{\(68.1\)} & \secondkey{\(9.6\)} & \secondkey{\(13.5\)} \\
\textbf{HAMoBE} & \ding{51} & \firstkey{\(54.3\)} & \firstkey{\(74.0\)} & \firstkey{\(15.1\)} & \firstkey{\(20.1\)} \\
\bottomrule
\end{tabular}
\caption{\small Ablation study on MEVID~\cite{davila2023mevid}.
\vspace{-3mm}}
\label{tab:moe_ablation}
\end{table}

%---------------------------------------------------------
\begin{figure*}[t]
  \centering
  \resizebox{0.9\linewidth}{!}{
  \includegraphics[trim=0 0 0 0,clip]{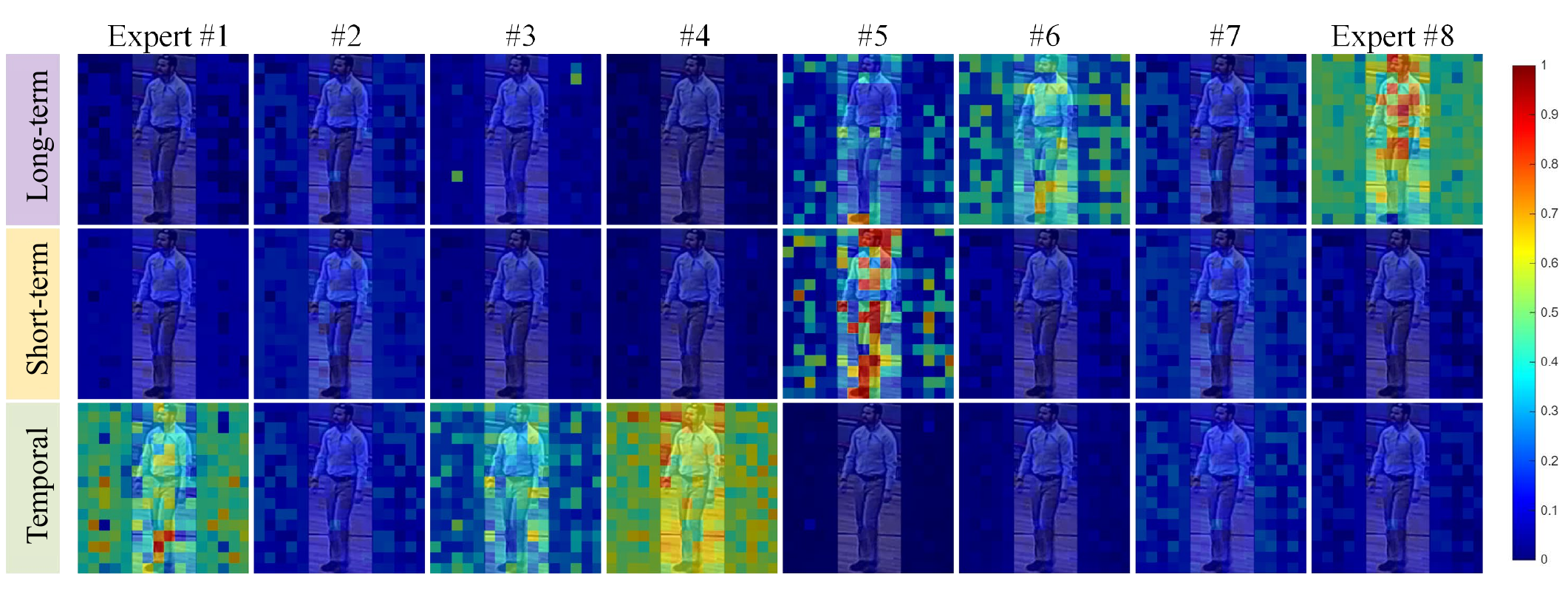}
  }
  \vspace{-2mm}
  \caption{\small Heatmaps of \(\mathbf{W}^{*1}\) illustrating the influence of eight first-layer experts on the long-term, short-term, and temporal features. Each row reflects an expert’s contribution. The CLS token is removed; remaining tokens are reshaped to $16 \times 16$ to match image tokens.
  \vspace{-1mm}}
  \label{fig:first_weight}
\end{figure*}
%---------------------------------------------------------

\Paragraph{Impact of Biometric Experts and Consistency Losses}
We evaluate the impact of our HAMoBE design and the incorporation of consistency losses by comparing with three baselines in Tab.~\ref{tab:moe_ablation}: a basic Vanilla MoE without consistency losses, our HAMoBE framework minus the consistency losses, and the full HAMoBE model with consistency losses.
The results indicate that the baseline Vanilla configuration achieves modest performance. Introducing the HAMoBE architecture improves accuracy, particularly under SC conditions, though improvements are less noticeable for DC. The full HAMoBE setup with consistency losses markedly improves outcomes across both clothing scenarios, emphasizing the role of both the MoE and consistency losses in boosting the model's discriminative power under varying conditions.

\Paragraph{Expert Contribution in First-Layer Gating Network}
Fig.~\ref{fig:first_weight} displays heatmaps on images, showing \(\mathbf{W}^{*1}\), how each of the $8$ first-layer experts contributes to features for long-term (\emph{i.e.}, static body shape), short-term (\emph{i.e.}, appearance), and temporal (\emph{i.e.}, pose/gait). Rows indicate different experts' influence on the biometric attributes. The color intensity in heatmaps shows the expert's impact—red signifies greater influence. This visualization highlights the varied focus of each expert and the complex feature interplay our model uses to improve re-identification accuracy.

\begin{figure}[t]
  \centering
  \resizebox{0.9\linewidth}{!}{
  \includegraphics[trim=0 0 0 0,clip]{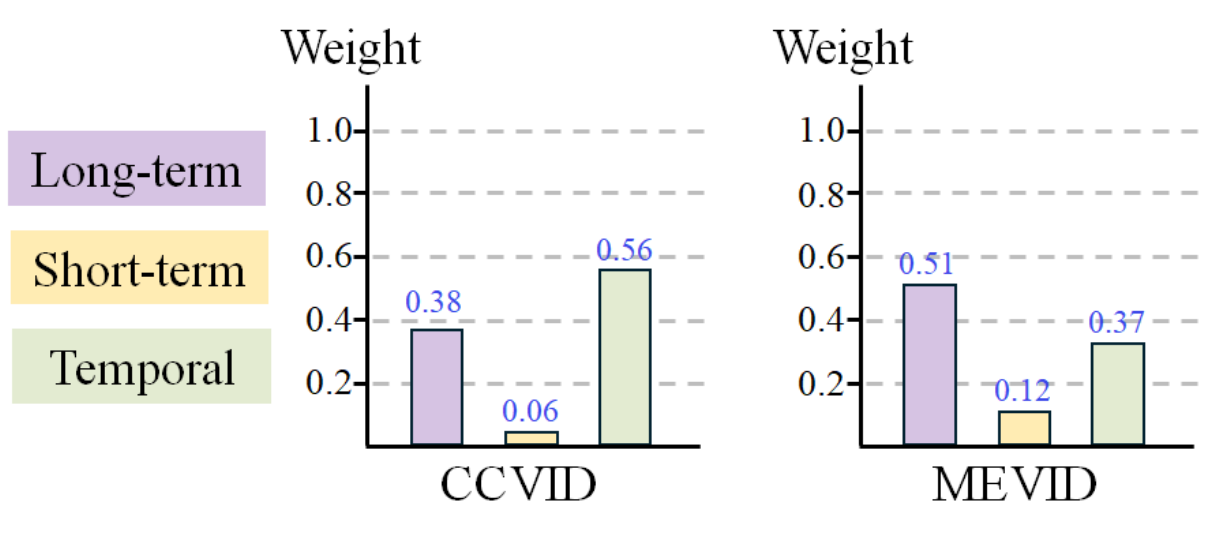}
  }
  \vspace{-3mm}
  \caption{\small Distribution of adaptive weights \({w}^{*2}\).
  \vspace{-3mm}}
  \label{fig:second_weight}
\end{figure}

\Paragraph{Adaptive Weight Distribution in Second-Layer Gating Network}
Fig.~\ref{fig:second_weight} illustrates the dynamic weights \({w}^{*2}\) from the second-layer gating network across two datasets: CCVID and MEVID.
For CCVID, the emphasis is predominantly on gait (0.56), whereas MEVID balances between appearance (0.51) and gait (0.37), reflecting the dataset's diverse conditions. This illustrates the model's adaptability in prioritizing biometric features to enhance person ReID across different datasets/scenarios.

\paragraph{Analysis on Dual-Input Mode.}
The single-input mode is faster, completing feature extraction in about $15$ minutes, but sacrifices some accuracy. 
In contrast, the dual-input mode takes  $\sim$$10$ minutes longer, yielding better accuracy (Tab.~\ref{tab:expert_ablation}). 
This can be further optimized by caching CLIP features, reducing redundancy and improving efficiency.

%---------------------------------------------------------
%---------------------------------------------------------

%% file: sec/5_conclusion.tex
\section{Conclusion}

In this work, we propose \textbf{HAMoBE}, a novel framework for video-based person ReID. 
%that capitalizes on the capabilities of the CLIP model for multi-scale feature extraction. 
By integrating biometric expertise at multiple layers—long-term, short-term, and temporal features—our approach dynamically and context-awarely fuses features to enhance identification accuracy. HAMoBE significantly surpasses existing methods in robustness and performance across diverse benchmarks. Its dual-input decision gating network further enhances adaptability, making it a highly effective solution for reID.

\paragraph{Acknowledgement}
This research is based upon work supported in part by the Office of the Director of National Intelligence (ODNI), Intelligence Advanced Research Projects Activity (IARPA), via 2022-21102100004. The views and conclusions contained herein are those of the authors and should not be interpreted as necessarily representing the official policies, either expressed or implied, of ODNI, IARPA, or the U.S. Government. The U.S. Government is authorized to reproduce and distribute reprints for governmental purposes notwithstanding any copyright annotation therein.

%% file: sec/X_supp.tex
\begin{figure}
  \centering
   %\fbox{\rule{0pt}{1in} \rule{0.9\linewidth}{0pt}}
   \includegraphics[width=1\linewidth]{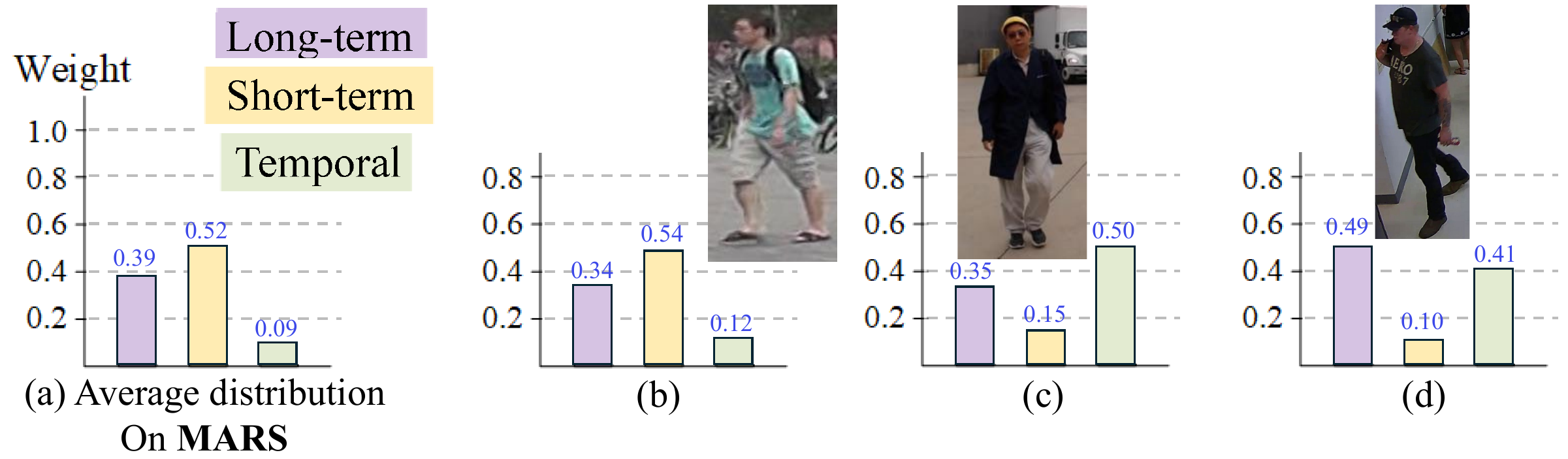}
  \vspace{-6mm}
  \caption{\small (a) Average \(\mathbf{W}^{*2}\) distribution on the MARS dataset. One example from (b) MARS, (c) CCVID, and (d) MEVID. }
  \label{fig:example}
\end{figure}

\section{Additional Experiments}

\Paragraph{Number of First-Layer Experts}
As suggested, we ablate the number of first-layer experts (4, 8, 16). Rank-1 accuracy under the different-clothes setting on CCVID improves from 82.6\% (4 experts) to \textbf{85.8\%} (8 experts), but drops to 84.5\% with 16, due to redundancy and reduced gating selectivity. This validates 8 as the most effective choice. 

\Paragraph{Ablaiton on Backbone}
We train a model using a ViT-B backbone to align with TF-CLIP. While this reduces overall capacity,
it still outperforms TF-CLIP in Rank-1 accuracy under the \emph{consistent-clothing setting}, achieving \textbf{90.5\%} (-0.6\%) vs. 89.4\% on MARS and \textbf{84.8\%} (-0.4\%) vs. 83.8\% on LS-VID. This highlights that our gains primarily stem from the framework design rather than the backbone size.

\Paragraph{Temporal Aggregation Strategy}  
We compare 4 temporal aggregation methods on MEVID under the different-clothes setting: mean pooling (9.8\%), LSTM (12.6\%), Transformer encoder (15.9\%), and our Transformer decoder (\textbf{20.1\%}). The decoder performs best, as its learnable query enables selective, frame-aware aggregation, offering better robustness to appearance changes.

\Paragraph{Additional Visualizations}
Fig.\ref{fig:example}{(a)} shows expert weight distributions in consistent-clothing scenarios. The increased contribution of \textbf{\emph{short-term}} experts aligns with our design, which emphasizes appearance cues when clothing is stable.

If we compare Fig.~\ref{fig:second_weight} and Fig.~\ref{fig:example}, expert weights adapt meaningfully to context: long-term \textbf{\emph{dominates}} on MEVID, short-term on MARS (\emph{consistent-clothing}), and temporal on CCVID. These shifts reflect the model’s ability to prioritize identity-relevant cues under varying conditions. Concrete examples in Fig.~\ref{fig:example} further illustrate distinct expert dominance across cases, ruling out fixed bias and supporting effective disentanglement.

\section{Limitations and Societal Impacts}
Despite advancements, HAMoBE has limitations that require further exploration. Currently, HAMoBE does not incorporate background information, potentially reducing its effectiveness in cluttered environments.
Its performance depends heavily on the quality and variability of input video data, making it less effective in low-resolution or occluded conditions. 
While HAMoBE aims to enhance person ReID for public safety, it also raises ethical concerns regarding privacy and surveillance, underscoring the need for stringent safeguards to prevent misuse.

\begin{figure*}
    \centering
    \includegraphics[width=\textwidth]{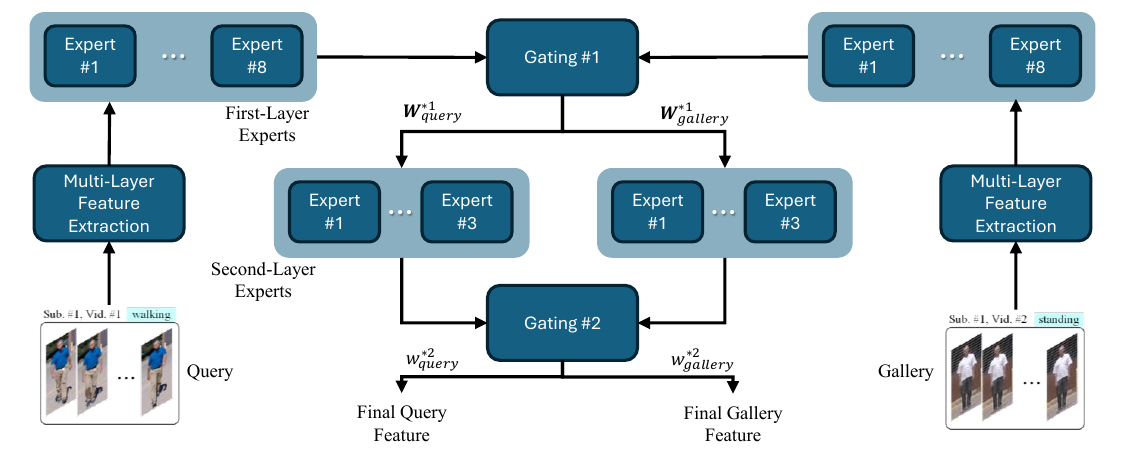}
    \caption{An illutration of the full model in dual-input mode.}
    \label{fig:flowchart_dual}
\end{figure*}